\documentclass[letterpaper, 10 pt, conference]{ieeeconf}  

\IEEEoverridecommandlockouts                              

\usepackage{graphicx}
\usepackage{bm}
\usepackage{cite}
\usepackage{subfig}
\usepackage{multirow}

\title{\LARGE \bf Accurate Contact Localization and Indentation Depth Prediction\\With an Optics-based Tactile Sensor 
}

\author{Pedro Piacenza$^{1}$, Weipeng Dang$^{2}$, Emily Hannigan$^{1}$ Jeremy Espinal$^{3}$, Ikram Hussain$^{3}$,\\ Ioannis Kymissis$^{2}$ and Matei Ciocarlie$^{1}$
\thanks{$^{1}$Department of Mechanical Engineering, Columbia University, New York, NY 10027, USA.}%
\thanks{\hspace{-3mm}{\tt\small \{pp2511,ejh2192,matei.ciocarlie\}@columbia.edu}}%
\thanks{$^{2}$Department of Electrical Engineering, Columbia University, New York, NY 10027, USA.}%
\thanks{\hspace{-3mm}{\tt\small wd2265@columbia.edu, johnkym@ee.columbia.edu}}%
\thanks{$^{3}$Columbia Engineering ENG Summer Research Program.}%
}

\begin{document}

\maketitle
\thispagestyle{empty}
\pagestyle{empty}
\setlength{\parskip}{0pt}

\begin{abstract}
Traditional methods to achieve high localization accuracy with tactile
sensors usually use a matrix of miniaturized individual sensors distributed on the area of interest. This approach usually
comes at a price of increased complexity in fabrication and circuitry,
and can be hard to adapt for non planar geometries. We propose to use
low cost optic components mounted on the edges of the sensing area to
measure how light traveling through an elastomer is affected by
touch. Multiple light emitters and receivers provide us with a rich
signal set that contains the necessary information to pinpoint both
the location and depth of an indentation with high accuracy. We
demonstrate sub-millimeter accuracy on location and depth on a 20mm by
20mm active sensing area. Our sensor provides high depth sensitivity
as a result of two different modalities in how light is guided through
our elastomer. This method results in a low cost, easy to manufacture sensor. We believe this approach can be adapted to cover
non-planar surfaces, simplifying future integration in robot skin
applications.

\end{abstract}

\section{Introduction}

Tactile sensors for robot manipulators can be analyzed and quantified
based on multiple performance criteria. From an operational
perspective, these include high accuracy in establishing both the
\textit{location} of a contact and the \textit {magnitude} of the applied force. In particular, good signal-to-noise ratio is
desirable for both the contact forces that characterize incipient
contact and the larger forces encountered during manipulation. From a
manufacturing perspective, achieving \textit{coverage} of potentially
irregular, non-flat surfaces is also important for application to
robotic fingers and palms.

One way to achieve accuracy and good coverage is by using individual
taxels distributed over the surface that must be sensorized; however,
this imposes miniaturization constraints on each taxel. Matrix
addressing for taxel arrays reduces the complexity of the circuit and
allows part of it to be implemented on a printed circuit board, but
imposes 2D structure on the sensor. As recent reviews point out,
achieving multiple such performance metrics traditionally leads to
manufacturing difficulties and system constraints that prohibit
large-scale deployment~\cite{DAHIYA10,yousef2011}.

\begin{figure}[t]
\centering
\includegraphics[width=0.65\linewidth]{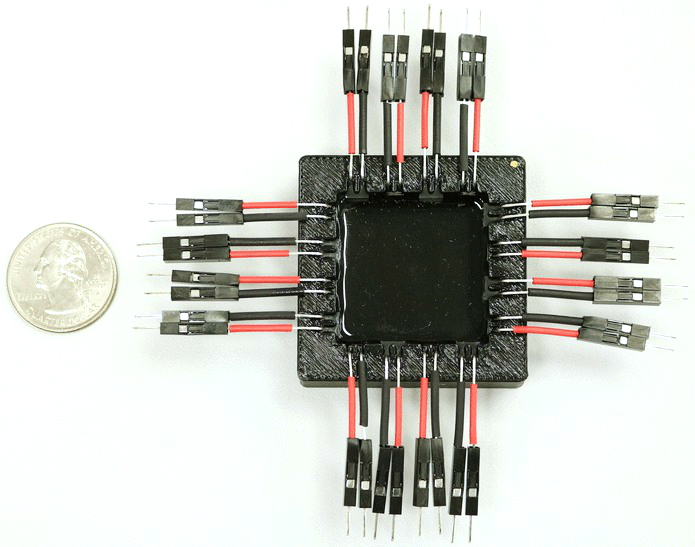}
\caption{Sensor consists of a square mold where LEDs and
  photodiodes are edge-mounted and the cavity is filled with an
  elastomer. We measure light transport through the elastomer
  to learn the location and the depth of an indentation.}
\label{fig:cover}
\end{figure}

Our approach is to build a tactile sensor as a continuous volume of a
transparent polymer, with light emitters and receivers embedded along
the perimeter (Fig.~\ref{fig:cover}). Indentation of the sensor area
affects how light from the emitters is transported through this
medium, producing a signal that is measured by the
receivers. Throughout this study, we use indentation depth as a proxy
for contact force, based on a known stiffness curve for the
constituent material. This method natively lends itself to covering
large areas with simple-to-manufacture, low-cost sensors.

In order to achieve the performance goals stated above, namely high
accuracy in both localization and indentation depth prediction, we
rely on two key ideas, which are also the main contributions of this
paper:
\begin{itemize}
\item \textit{Leverage multiple modes of interaction between an
  indenting probe and a light-transporting medium in order to increase
  depth prediction accuracy throughout the operating range.} As we
  will detail in the paper, indentation of the medium affects light
  transport in at least two ways. First, during initial contact, the
  probe alters the geometry of the surface and changes the refraction
  of the light rays. Second, as indentation becomes deeper, the probe
  blocks direct paths between emitters and receivers. We design our
  sensor to use both modes in continuous fashion, resulting in a good
  sensitivity throughout a wide range of indentation depths.
\item \textit{Use data-driven methods to directly learn the mapping
  between a rich signal set extracted from the sensor and our
  variables of interest.} For our sensor, such data set can be obtained by measuring the signal between every
  emitter and every receiver. In the past, we have used an all-pairs
  approach on a piezoresistance-based sensor and showed it can lead to
  high localization accuracy~\cite{PIACENZA16}.
\end{itemize}

\noindent
The result of using these methods is a tactile pad that exhibits desirable performance characteristics in
accuracy and sensitivity, while using a simple manufacturing method
and low-cost components. While not explicitly tested here, we believe
that both the fabrication technique and the data-driven signal
processing approach also lend themselves to constructing pads of
irregular three-dimensional geometry. Furthermore, mapping the signal
between every emitter and receiver produces a rich signal set with
relatively few wires. Both of these characteristics could enable
easier integration into robotic fingers and palms,
which is our directional goal.

\section{Related Work}

The use of optics for tactile sensing is not new, and has a long
history of integration in robotic fingers and hands. Early work by
Begej demonstrated the use of CCD sensors recording light patterns
through a robotic tip affected by deformation~\cite{BEGEJ88}. More
recently Lepora and Ward-Cherrier~\cite{LEPORA151} showed how to
achieve super-resolution and hyperacuity with a CCD-based touch sensors
integrated into a fingertip. Johnson and Adelson used color-coded 3D
geometry reconstruction to retrieve minute surface details with an
optics-based sensor~\cite{schneider2009}. These studies share a common
concept of a CCD array imaging a deformed fingertip from the inside,
requiring that the array be positioned far enough from the surface in
order to image the entire touch area. In our approach, the sensing
elements are fully distributed, allowing for coverage of large areas
and potentially irregular geometry. Work by Polygerinos et al.~\cite{polygerinos2010} use the deformation of an optic fiber to create a force transducer. This approach has the advantage that the sensing
electronics do no have to be located close to the contact area.

In our work, we take advantage of multiple modes of light transport
through an elastomer to
increase the sensitivity of the sensor. In recent work, Patel and
Correll took advantage of reflection and refraction to build an IR
touch sensor that also functions as a proximity sensor ~\cite{PATEL16}. Their work however does not provide means to also localize contact.

We perform contact localization by combining signals from multiple
emitter-receiver pairs, a technique which we previously used in the
context of a piezoresistive sensor~\cite{PIACENZA16}. Other sensors also use a small number of
underlying transducers to recover richer information about the
contact. For example, work in the ROBOSKIN project showed how to
calibrate multiple piezocapacitive transducers~\cite{cannata2010}, used
them to recover a complete contact profile~\cite{muscari2013} using an
analytic model of deformation, and finally used such information for
manipulation learning tasks~\cite{argall2011}. Our localization method
is entirely data driven and makes no assumptions about the underlying
properties of the medium, which could allow coverage of more complex
geometric surfaces.

Our localization approach shares some of the same goals of techniques
such as super-resolution and electric impedance tomography. Van den
Heever et al. ~\cite{HEEVER09} used a similar algorithm to
super-resolution imaging, combining several measurements of a 5 by 5
force sensitive resistors array into an overall higher resolution
measurement. Lepora and Ward-Cherrier\cite{LEPORA151} and Lepora et
al.\cite{LEPORA152} used a Bayesian perception method to obtain a
35-fold improvement of localization acuity (0.12mm) over a sensor
resolution of 4mm. Electric impedance tomography (EIT) is used to
estimate the internal conductivity of an electrically conductive body
by virtue of measurements taken with electrodes placed on the boundary
of said body. While originally used for medical applications, EIT
techniques have been applied successfully for manufacturing artificial
sensitive skin for robotics~\cite{NAGAKUBO07,KATO07,TAWIL11}, although 
with lower spatial resolution than other methods.

We rely on data-driven methods to learn the behavior of our sensors;
along these lines, we note that machine learning for manipulation
based on tactile data is not new. Ponce Wong et al.~\cite{PONCE14}
learned to discriminate between different types of geometric features
based on the signals provided by a previously
developed~\cite{WETTELS08} multimodal touch sensor. Current work by
Wan et al.~\cite{WAN16} relates tactile signal variability and
predictability to grasp stability using recently developed MEMS-based
sensors~\cite{TENZER14}. With traditional tactile arrays, Dang and
Allen~\cite{ALLEN13} successfully used an SVM classifier to
distinguish stable from unstable grasps in the context of robotic
manipulation using a Barrett Hand. Bekiroglu et al.~\cite{BEKIROGLU11}
also studied how grasp stability can be assessed based on tactile
sensory data using machine-learning techniques. In similar fashion, both Saal et
al. \cite{SAAL10} and Tanaka et al.~\cite{TANAKA14} used probabilistic
models on tactile data to estimate object dynamics and perform object
recognition respectively. However, most of this work is based on
arrays built on rigid substrates and thus unable to provide full
coverage of complex geometry. In contrast, we apply our methods to the
design of the sensor itself, and believe that developing the sensor
simultaneously with the learning techniques that make use of the data
can bring us closer to achieving complete tactile systems.

\section{Tactile Sensing Method}

\begin{figure}[ht]
\centering
\begin{tabular}{c}

\includegraphics[clip, trim=1.5cm 3cm 1.5cm 3.5cm, width=0.70\linewidth]{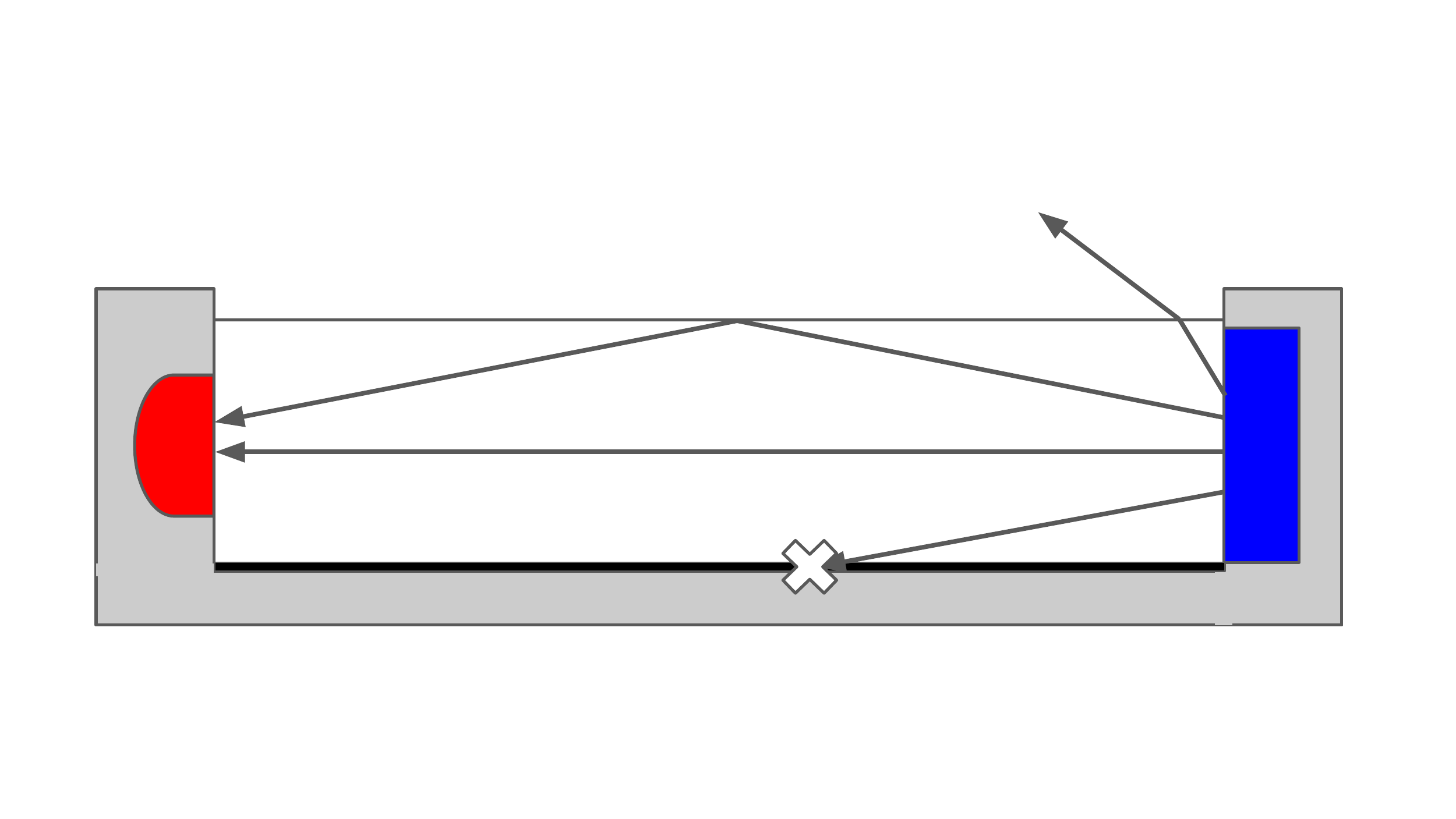}\\
{\footnotesize a) Sensor Undisturbed}\\[3mm]

\includegraphics[clip, trim=1.5cm 3cm 1.5cm 3.5cm, width=0.70\linewidth]{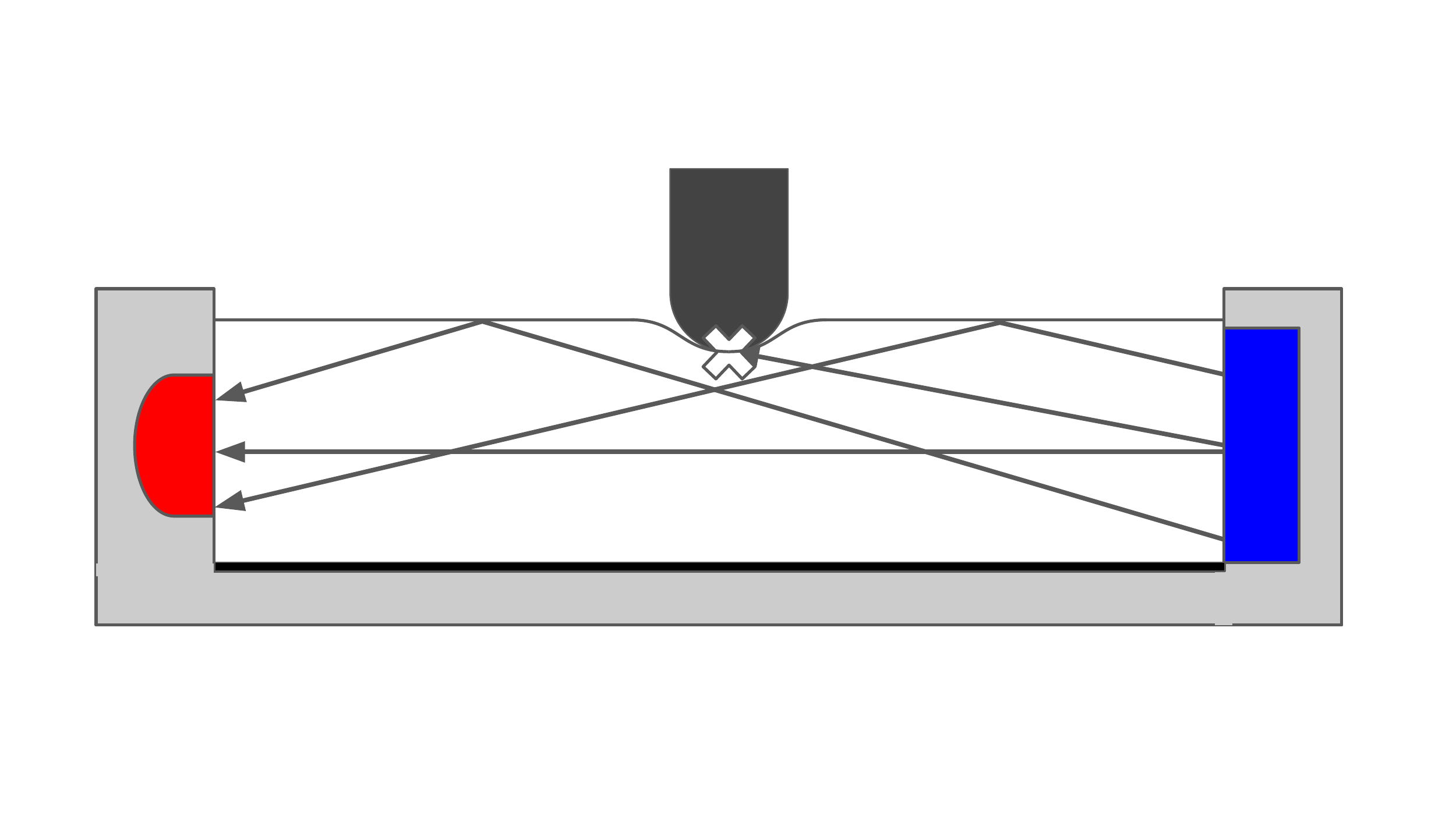}\\
{\footnotesize b) First mode of detection}\\[3mm]

\includegraphics[clip, trim=1.5cm 3cm 1.5cm 3.5cm, width=0.70\linewidth]{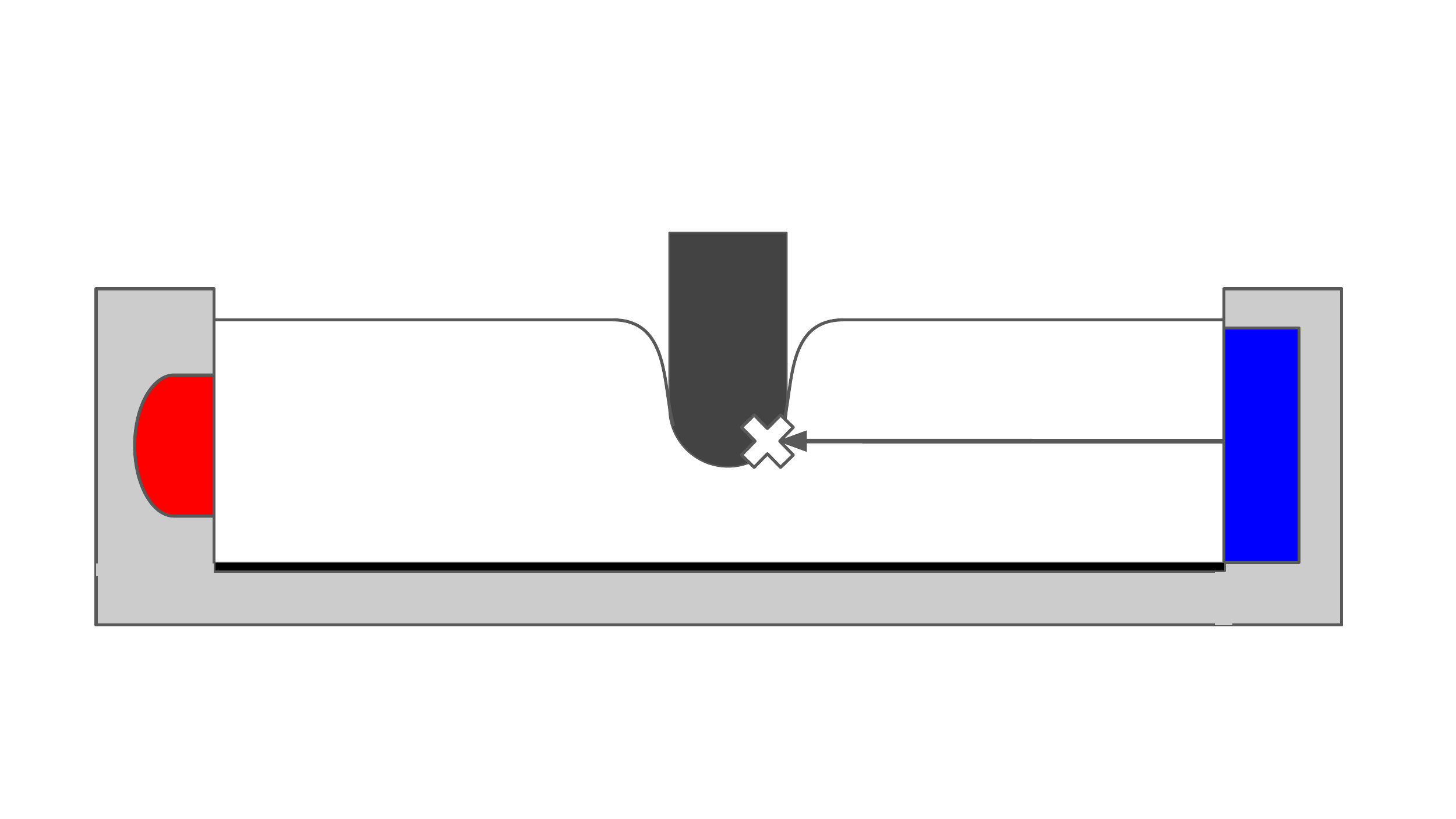}\\
{\footnotesize c) Second mode of detection}

\end{tabular}
\caption{The first mode of detection ($b$) is the result of light
  scattering and surface deformation. The second mode ($c$) is the
  result of the indenter tip physically blocking the direct path of
  light.}
\label{fig:Modes}
\vspace{-0mm}
\end{figure}

The fundamental sensing unit of our approach is comprised of a light
emitting diode (LED) and a photodiode receiver, edge-mounted around a
sensing area which is filled with polydimethylsiloxane (PDMS)
(Fig.~\ref{fig:Modes}a). While we ultimately use sensors with
multiple emitters and receivers, in this section we focus on a single
emitter-receiver pair in order to discuss the underlying transduction
mechanism; we will return to complete sensor design in the next
section.

\subsection{Light transport and interaction modes}

\begin{figure}[t]
\centering
\begin{tabular}{c}

\includegraphics[clip, trim=0.4cm 0.5cm 0.5cm 0.5cm, width=0.8\linewidth]{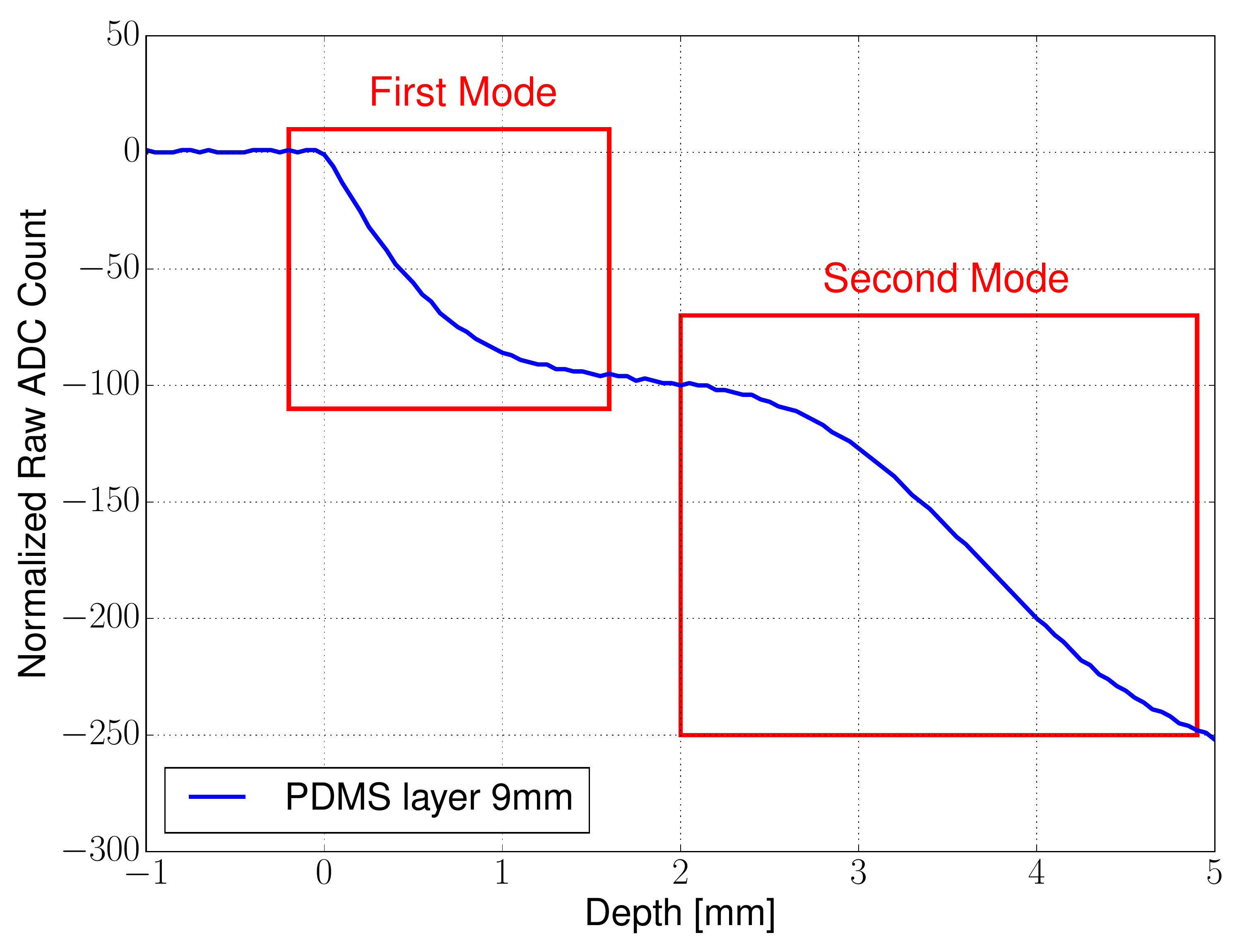}\\
\\
\includegraphics[clip, trim=0.4cm 0.5cm 0.5cm 0.5cm, width=0.8\linewidth]{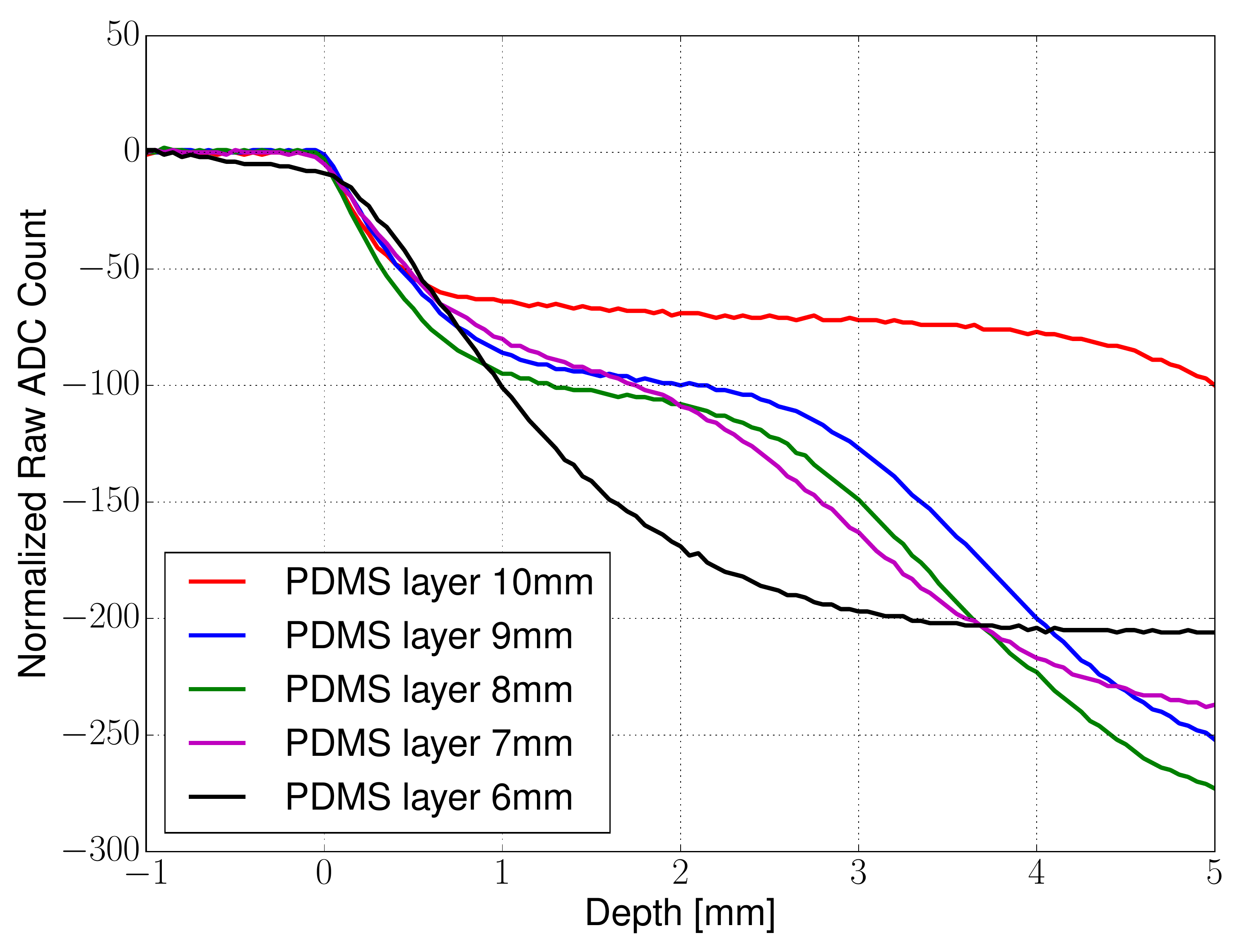}\\
\end{tabular}

\caption{The first mode of detection happens upon light contact and manifests as a sudden drop in the signal. The second mode is activated with a heavier contact, in this particular case after a depth of approximately 2mm}
\label{fig:Signal_modes}
\end{figure}

The core transduction mechanism relies on the fact that as a probe
indents the surface of the sensor, light transport between the emitter
and the receiver is altered, changing the signal reported by the
receiver. Consider the multiple ways in
which light from the LED can reach the opposite photodiode: through a
direct path or through a reflection. Of particular interest to us are
reflections off the surface of the sensor: based on Snell's law, due
to different refractive indices of the elastomer and air, light rays
hitting the surface below the critical angle are reflected back into
the elastomer.

As the probe makes initial contact with the sensor surface, the
elastomer-air interface is removed from the area of contact;
furthermore, surface normals are immediately disturbed. This changes
the amount of light that can reach the diode via surface reflection
(Fig.~\ref{fig:Modes}b). \textit{This is the first mode of interaction
  that our transduction method captures.} It is highly sensitive to
initial contact, and requires very little penetration depth to produce
a strong output signal.

As the depth of indentation increases, the object penetration into the
PDMS also starts to block the light rays that were reaching the
photodiode through a direct, line-of-sight path
(Fig.~\ref{fig:Modes}c). \textit{This is our second mode of
  interaction.} To produce a strong signal, the probe must reach deep
enough under the surface where it blocks a significant part of the
diode's surface from the LED's vantage point.

We note that other light paths are also possible between the emitter
and receiver. The interface between the clear elastomer
and the holding structure (the bottom and side walls of the cavity)
can also give rise to reflections. In practice, we have found
that the elastomer and the holding plastic exhibit bonding/unbonding
effects at a time scale of 5-10s when indented, creating unwanted hysteresis. 

To eliminate such effects, we coat the bottom of the sensor with a
1mm layer of elastomer saturated with carbon black particles
(shown in Fig.~\ref{fig:Modes} by a thick black line). This eliminates bottom surface reflections, and exhibits no
adverse effects, as the clear elastomer permanently
bonds with the carbon black-filled layer. There are still possible
reflections off the walls of the sensor; while we do not explicitly
consider their effects, they can still produce meaningful signals
that are used by data-driven mapping algorithm.

\subsection{Effective operating range and prototype construction}

We would like our sensor to take advantage of both operating modes
described above, noting that one is highly sensitive to small
indentations while the other provides a strong response to deeper
probes. \textit{We thus aim to design our sensor such that these two
  modes are contiguous as the indentation depth increases.} The goal is to obtain high sensitivity throughout the
operating range of the sensor. The key geometric factor affecting this
behavior is the height of the elastomer layer, which we determine
experimentally.

We constructed multiple 3D printed (black ABS material) molds with
LEDs (SunLED model XSCWD23MB) placed 20mm away from the photodiode
(Osram SFH 206K) on opposing sides of a square mold. These sensors
have 3 LEDs on one side of the mold, and 3 corresponding photodiodes
directly in front of the LEDs. The mold was filled with a transparent
elastomer (PDMS, Sylgard 184, Dow Corning). PDMS and air have
approximate refractive indexes of 1.4 and 1.0~\cite{johnson09}
respectively, leading to a critical angle of $45$ degrees. This means
that if light hits the boundary between PDMS and air at an angle
greater than $45$ degrees with respect to the surface normal, the ray
is reflected back into the PDMS where it can be detected by the
photodiode.

\begin{figure}[t]
\centering
\includegraphics[clip, trim=0.3cm 0.3cm 0.3cm 0.3cm, width=0.65\linewidth]{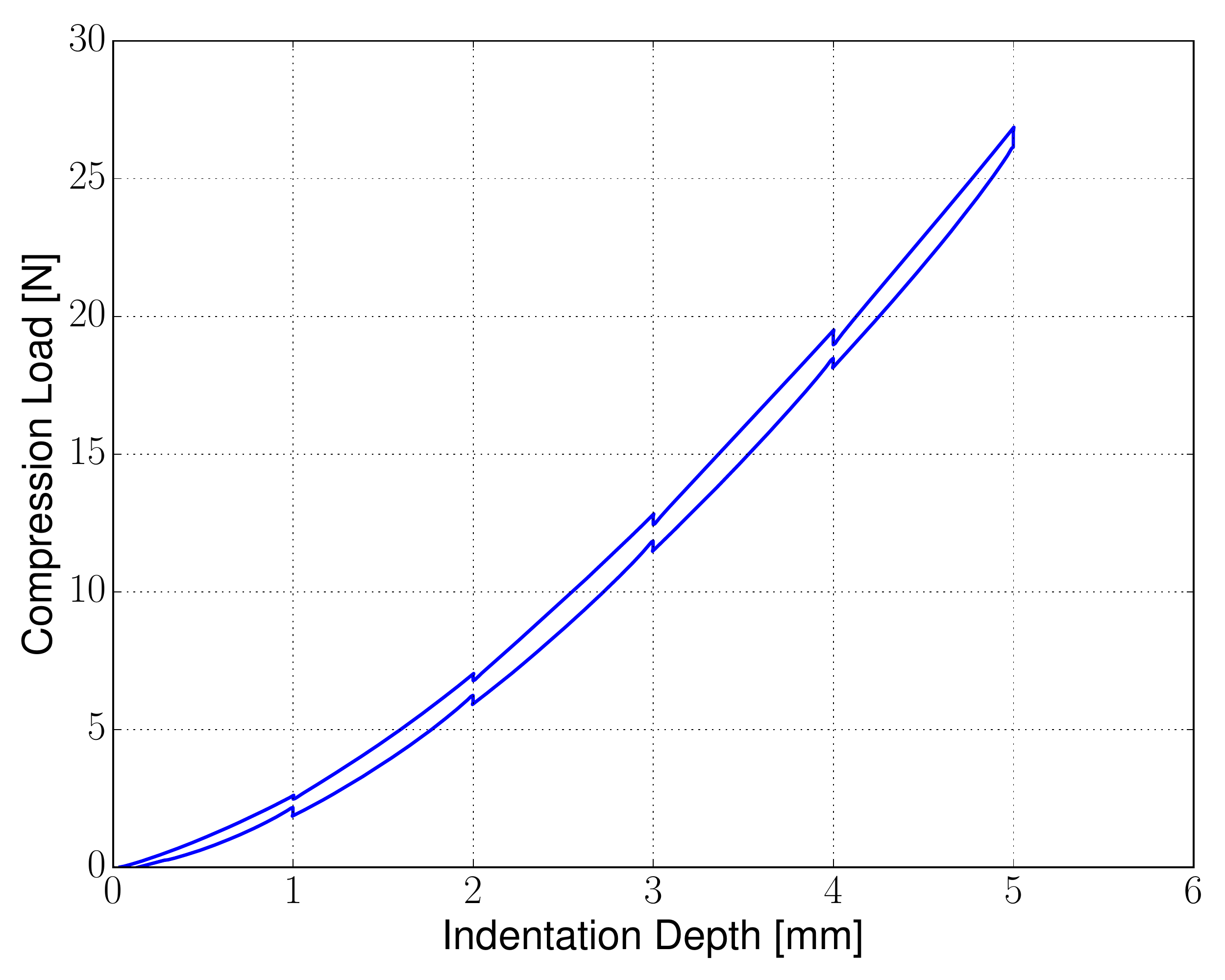}
\caption{Load vs indentation depth for a 1:20 ratio of curing agent to PDMS. Measurements were taken by advancing or retracting the probe in 1mm steps separated by 10s pauses}
\label{fig:Mapping}
\end{figure}

Prototypes with PDMS layers over 10mm
showed the presence of a deadband: after a
certain threshold depth the photodiode signal does not change as we
indent further down until you indent deep enough to activate the
second mode. To make these two modes continuous we build a set of
sensors where we vary the thickness of the PDMS
layer. Results can be visualized in Fig.~\ref{fig:Signal_modes}. Based on these results, while the 7mm layer provides the best
continuity between our two modes, the 8mm layer gives good continuity
while also producing a slightly stronger signal when
indented. We build our subsequent sensor to have an
8mm PDMS layer.

Another parameter to choose when building the sensor is the stiffness
of the PDMS. This parameter lets us directly manipulate the mapping
between indentation depth and indentation force. The stiffness of PDMS
is determined by the weight ratio between the curing agent and the
polymer itself. Fig.~\ref{fig:Signal_modes} corresponds to a ratio
of 1:20 curing agent to PDMS, which we used in our sensors. Fig.~\ref{fig:Mapping} shows the mapping between indentation depth up to
5mm and force for the 6mm hemispherical tip used in our experiments.

\section{Complete Sensor Design}

\begin{figure}[ht]
\centering \includegraphics[clip, trim=3.2cm 1cm 3.2cm 1.2cm,
  width=0.72\linewidth]{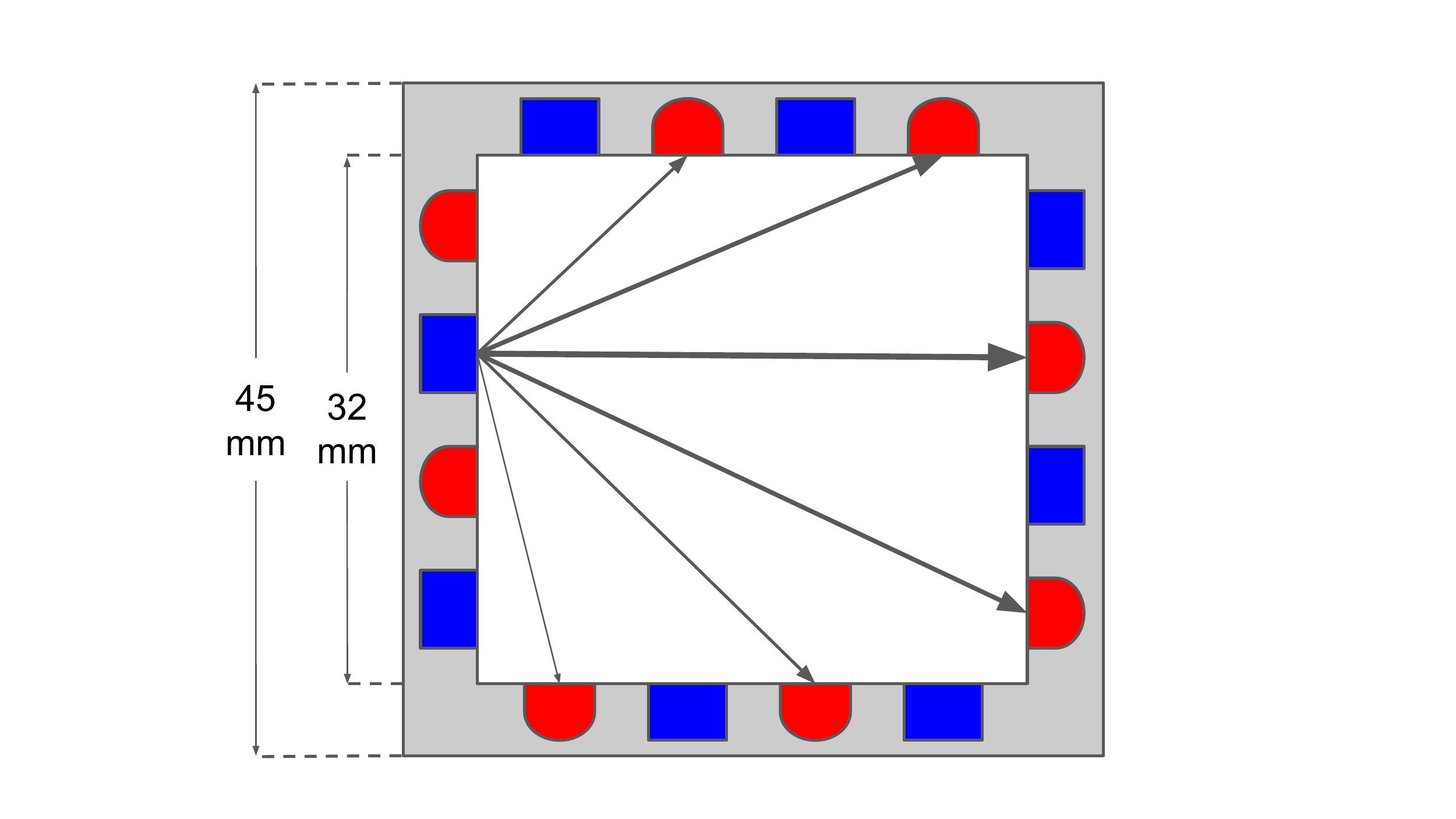}
\caption{The sensor design enables several receivers to be excited by a single emitter. This one-to-many relationship provides a rich set of signals when switching through all of our emitters and
  measuring the signals on all the receptors}
\label{fig:Sensor}
\end{figure}

While sensitivity to a large range of indentation depths (and forces)
is important for applications in manipulation, it is not
sufficient. The ability to localize touch accurately on a 2D surface
embedded in 3D space is also
critical. To achieve this goal, we construct our sensors with numerous light emitters and receivers mounted around the
perimeter (Fig.~\ref{fig:Sensor}). This gives rise to numerous
emitter-receiver pairs, each behaving as a unit described in the
previous section.

The multi-pair approach gives us a very rich signal set, with
cardinality equal to the number of emitters multiplied by the number
of receivers. We make the assumption that an indentation anywhere on
the sensor will affect multiple such signals. We then use a
data-driven approach to directly learn how these signals map to our
variables of interest, such as indentation location and depth. 

We validate this concept on a sensor comprised of 8 LEDs and 8
photodiodes arranged in an alternating pattern and mounted in sockets along the central cavity walls. To build the sensor we use a 3D printed square mold with exterior dimensions of 45mm x 45mm. The cavity in the mold is 32mm x 32mm. Light travels from emitters to receivers via multiple paths that cover the sensing area. This way any LED is able to excite several
photodiodes (Fig.~\ref{fig:Sensor}).

To determine the location
and depth of the indentation, we will read signals from all the
photodiodes as different LEDs turn on. Having 8 LEDs gives rise to 8
signals for each photodiode; plus an additional signal with all LEDs
turned off. This last signal allows us to measure the ambient light captured by each diode, and leads to a total of 9 signals per diode. One important consideration with a sensor where the sensing units are exposed to ambient light is to incorporate this information such that the sensor can perform
consistently in different lighting situations. To achieve this, we use
the state where all LEDs are OFF as a baseline that gets removed from
every other state at each sampling time.

An Arduino Mega 2560 handles switching between our 9 states, and
taking analogue readings of each photodiode. The photodiode signal is
amplified through a standard transimpedance amplifier circuit, and
each LED is driven at full power using an NPN bipolar junction
transistor. The resulting sampling frequency with this setup is
60Hz.

\section{Data Collection Protocol}

Data collection is performed using a planar stage
(Marzhauser LStep) and a linear probe located above the stage to
indent vertically on the sensor with a 6mm hemispherical tip. The
linear probe is position-controlled and the
reference level is set manually such that the indenter
tip barely makes contact with the sensor surface. The linear probe does not
have force sensing capability, hence we use indentation depth
as a proxy for indentation force.

Two patterns are used to indent our sensor. The
\textit{grid indentation pattern} indents the sensor on a 2mm regular
grid. Taking into account the diameter of our tip, plus a 3mm margin
such that we don't indent directly next to an edge, this results in 121 indent locations distributed over a 20mm x 20mm area. The order of
indentation within this grid is randomized. The \textit{random
  indentation pattern} indents the sensor in randomly generated
locations within its workspace.

At each location we follow the same protocol. Consider the sensor
surface to be the reference level, and positive depth values
correspond to the indenter tip going deeper into the sensor. To discriminate touch vs. non-touch conditions, we
collect data at both negative and positive depths. For depths between
$-10mm$ and $-1mm$, we collect one data point every $1mm$. The
indenter then goes down to a depth of $5.0mm$ taking measurements
every $0.1mm$. The same procedure is mirrored with the indenter tip
retracting.

Each measurement $i$ results in a tuple of the form
$\Phi_i=(x_i,y_i,d_i,p_{j=1}^1,..,p_{j=1}^8,...,p_{j=9}^1,..,p_{j=9}^8)$
where $(x_i,y_i)$ is the indentation location in sensor coordinates,
$d_i$ is the depth at which the measurement was taken and
$(p_{j}^1,..,p_{j}^8)$ corresponds to the readings of our 8
photodiodes at state $j\ \epsilon\ [1,9]$. For each diode, states 1
through 8 correspond to each one of the 8 LEDs being ON, and state 9
corresponds to the case where all the LEDs are OFF. We thus have a
total of 75 numbers comprised in each tuple $\Phi_i$. These tuples are
analyzed as described in the following section.

\section{Analysis and Results}

\begin{figure*}[t]

\subfloat[Location (16.5, 19.1)]{\label{dex0}
\includegraphics[width=0.28\textwidth]{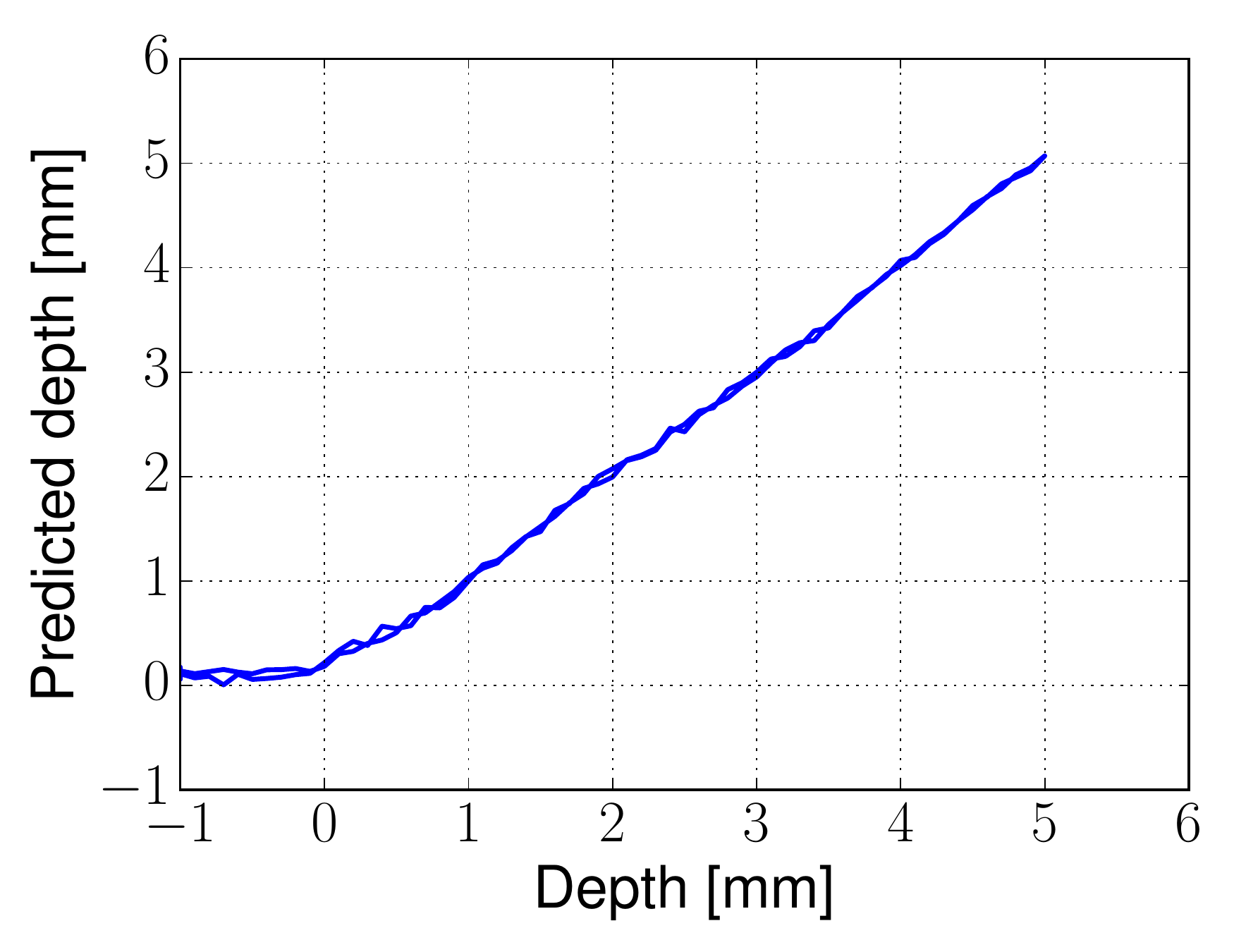}}\hfill
\subfloat[Location (6.4, 18.3)]{\label{dex1}
\includegraphics[width=0.28\textwidth]{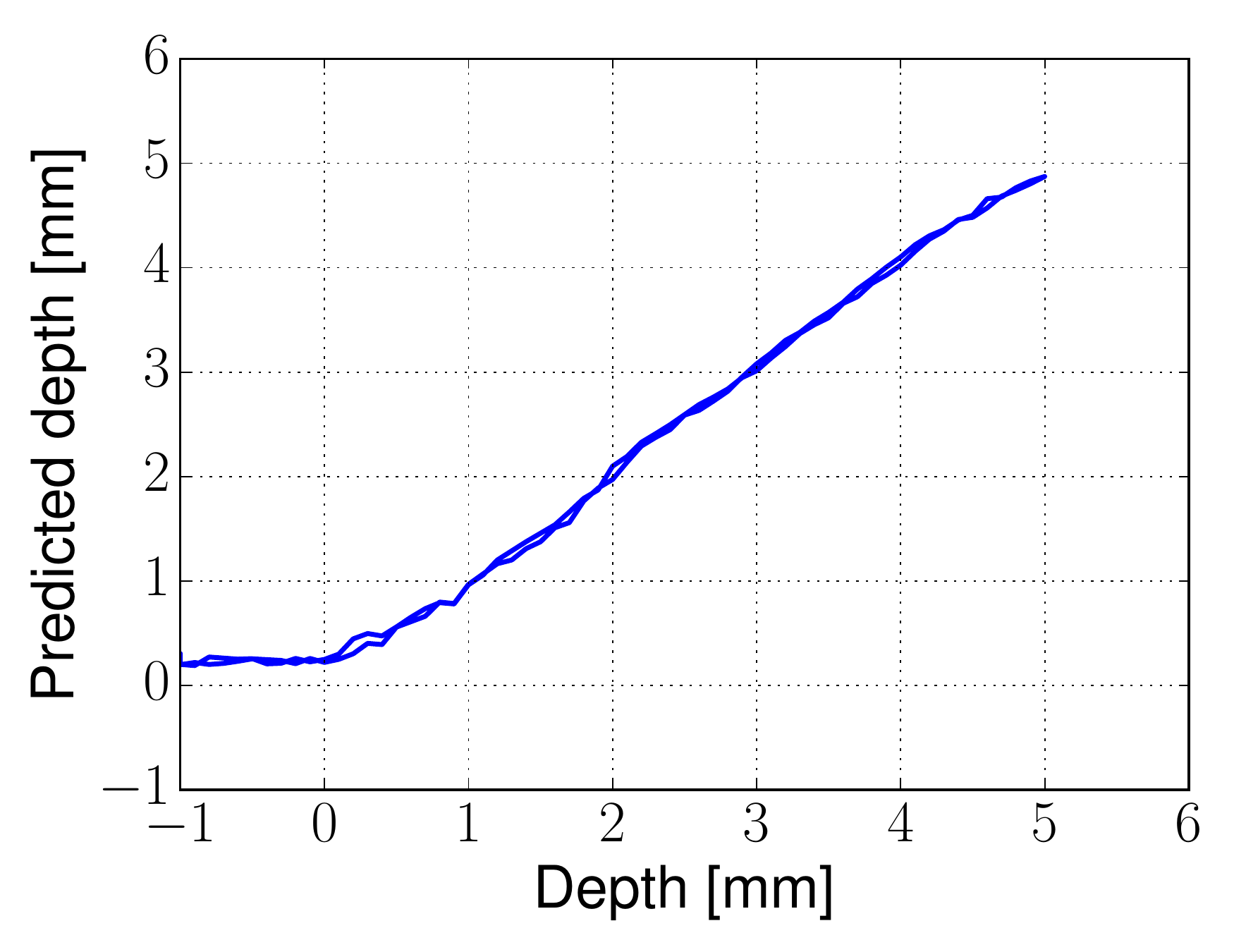}}\hfill
\subfloat[Location (10.9, 6.2)]{\label{dex2}
\includegraphics[width=0.28\textwidth]{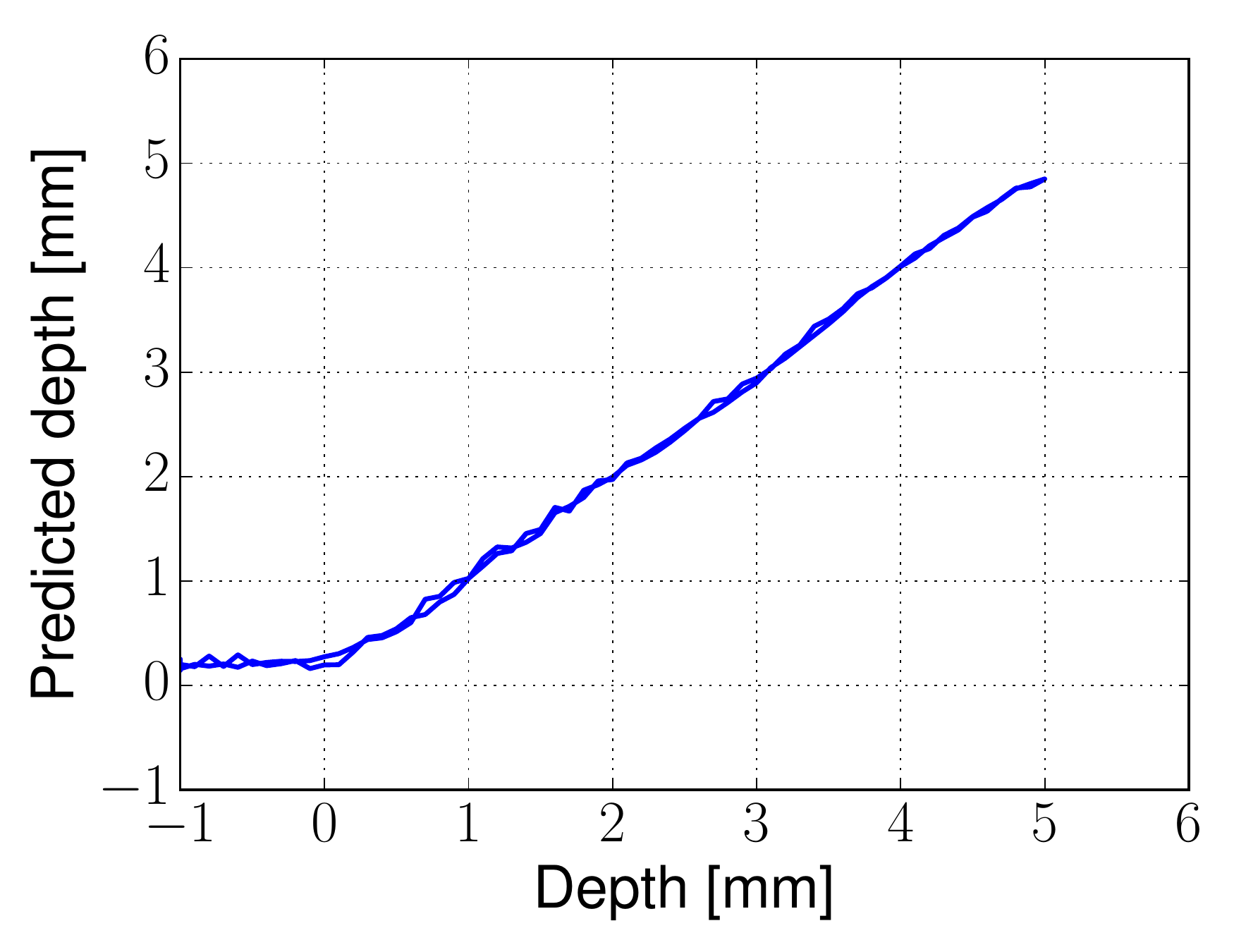}}\\

\caption{ Regression results for depth prediction at a central location ($a$), an edge location close to the x-axis ($b$) and another edge location close to the y-axis of the sensor($c$). These are some of the random locations in our test dataset }
\label{fig:depth}
\end{figure*}

Our main objective is to learn the mapping between our photodiode
readings $(p_{j}^1,..,p_{j}^8)$ to the indentation location and depth
$(x_i,y_i,d_i)$. 

We have found that we obtain higher performance by splitting this
problem into two components. First, we use a classifier to determine
if touch is occurring; this classifier is trained on both data points
with $d_i < 0$ and $d_i \geq 0$. We use a linear SVM as our classifier
of choice for this problem.

If the SVM classifier predicts touch is occurring, we use a second
stage regressor that outputs predicted values for
$(x_i,y_i,d_i)$. This regressor is trained only on training data with
$d_i \geq 0$. We use a kernelized ridge regressor with a Laplacian
kernel and use half of the training data to calibrate the
ridge regression tuning factor $\lambda$ and the kernel bandwidth
$\gamma$ through grid search. Results presented in this section were
obtained with $\lambda = 2.15e^-4$ and $\gamma = 5.45e^-4$.

\begin{table}[b]
\centering
\caption{Touch vs. no touch classification success rate}
\label{Depth_table}
\begin{tabular}{cccc}
\hline
\\[-3mm]
Ground Truth & Depth Value & \begin{tabular}[c]{@{}c@{}}Ambient Light\\ test set\end{tabular} & \begin{tabular}[c]{@{}c@{}}Dark test\\ set\end{tabular} \\ \hline
\\[-3mm] \hline
\\[-2mm]
No touch     & -0.4 mm     & 1.0                                                              & 1.0                                                     \\
No touch     & -0.2 mm     & 0.99                                                             & 1.0                                                     \\
Touch        & 0 mm        & 0.04                                                             & 0.10                                                    \\
Touch        & 0.2 mm      & 0.32                                                             & 0.33                                                    \\
Touch        & 0.4 mm      & 0.51                                                             & 0.65                                                    \\
Touch        & 0.6 mm      & 0.76                                                             & 0.90                                                    \\
Touch        & 0.8 mm      & 0.96                                                             & 0.96                                                    \\
Touch        & 1.0 mm      & 0.98                                                             & 0.98                                                   
\end{tabular}
\end{table}

To train our predictors we collected four grid pattern datasets, each
consisting of 121 indentations, and each indentation containing 161
datapoints at different depths. Aiming for robustness to changes in
lighting conditions, two of these datasets were collected with the
sensor exposed to ambient light and the other two datasets were
collected in darkness. The feature space used for training has
a dimensionality of 64, since the all LEDs OFF state is first
subtracted from all other signals and not used as a stand-alone
feature.

The metric used to quantify the success of our regressor is
the magnitude of the error for both the localization and depth
accuracy. In case of the classifier, the metric is the percentage of
successful predictions. Each test dataset is collected on a
\textit{random indentation pattern} which contains 100 indentation
events. The results presented here are those obtained by testing our
models against two different test datasets: one collected in ambient
light and another collected in the dark.

Classification results in the region of interest for the touch vs
no-touch case are summarized in Table \ref{Depth_table}. Note that
these results are sliced at a certain depth, and they aggregate the
classification performance across all locations. With both test
datasets, the classifier has difficulty detecting touch at 0mm, where
the tip of the indenter is barely making contact with the
sensor. However, at 0.6mm depth the dark test dataset already provides a
90\% success rate in the classification, while the ambient light test
dataset provides 76\% success rate. According to the mapping presented on Fig.~\ref{fig:Mapping}, a depth of 0.6mm corresponds to an indentation
force of approximately 2 Newtons. This represents the minimum indenter force our sensor is capable of detecting.

\begin{figure*}[ph]

\subfloat[]{\label{ex0}
\includegraphics[clip, trim=0.25cm 1.2cm 0.25cm 0.25cm, width=0.30\textwidth]{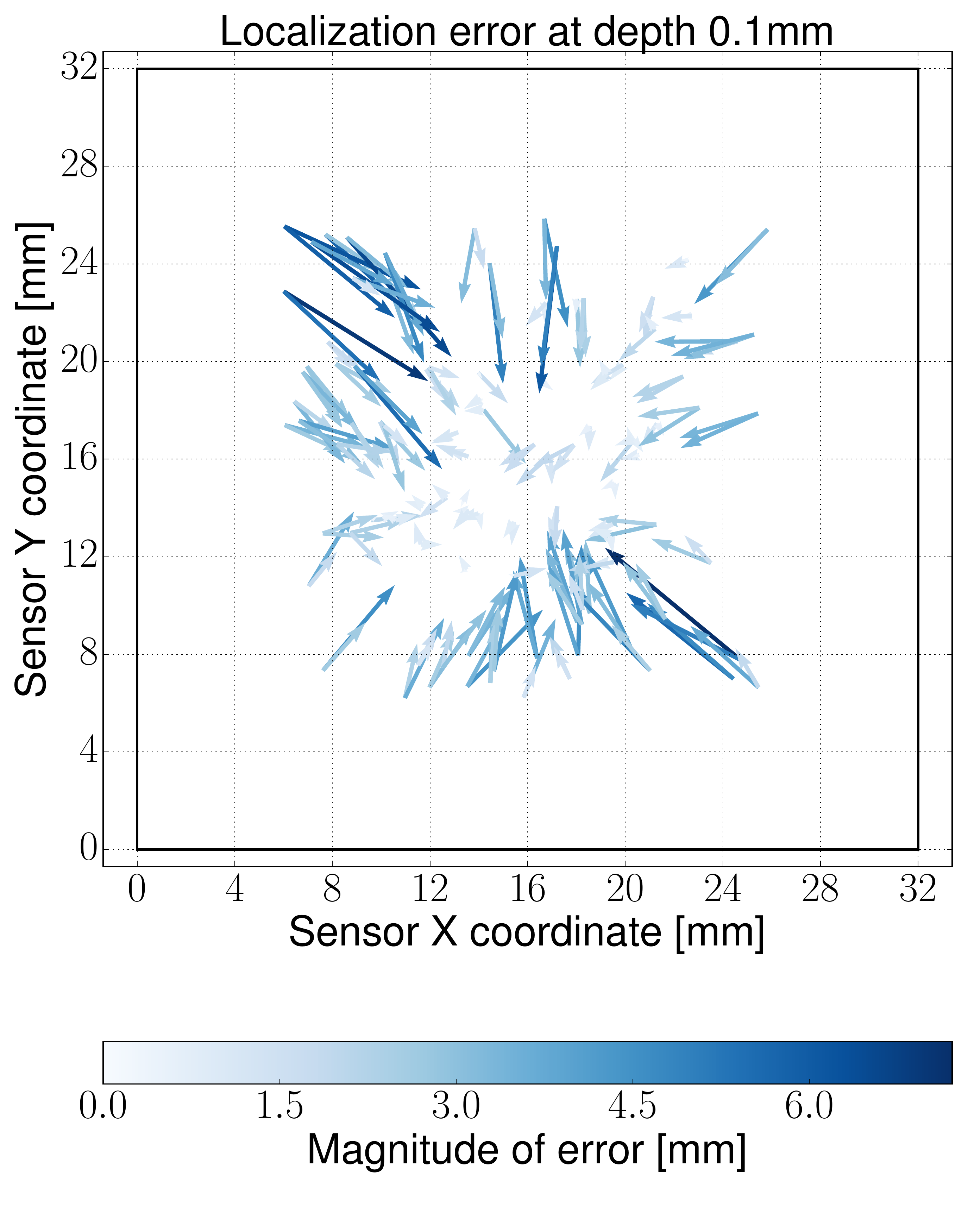}}\hfill
\subfloat[]{\label{ex1}
\includegraphics[clip, trim=0.25cm 1.2cm 0.25cm 0.25cm, width=0.30\textwidth]{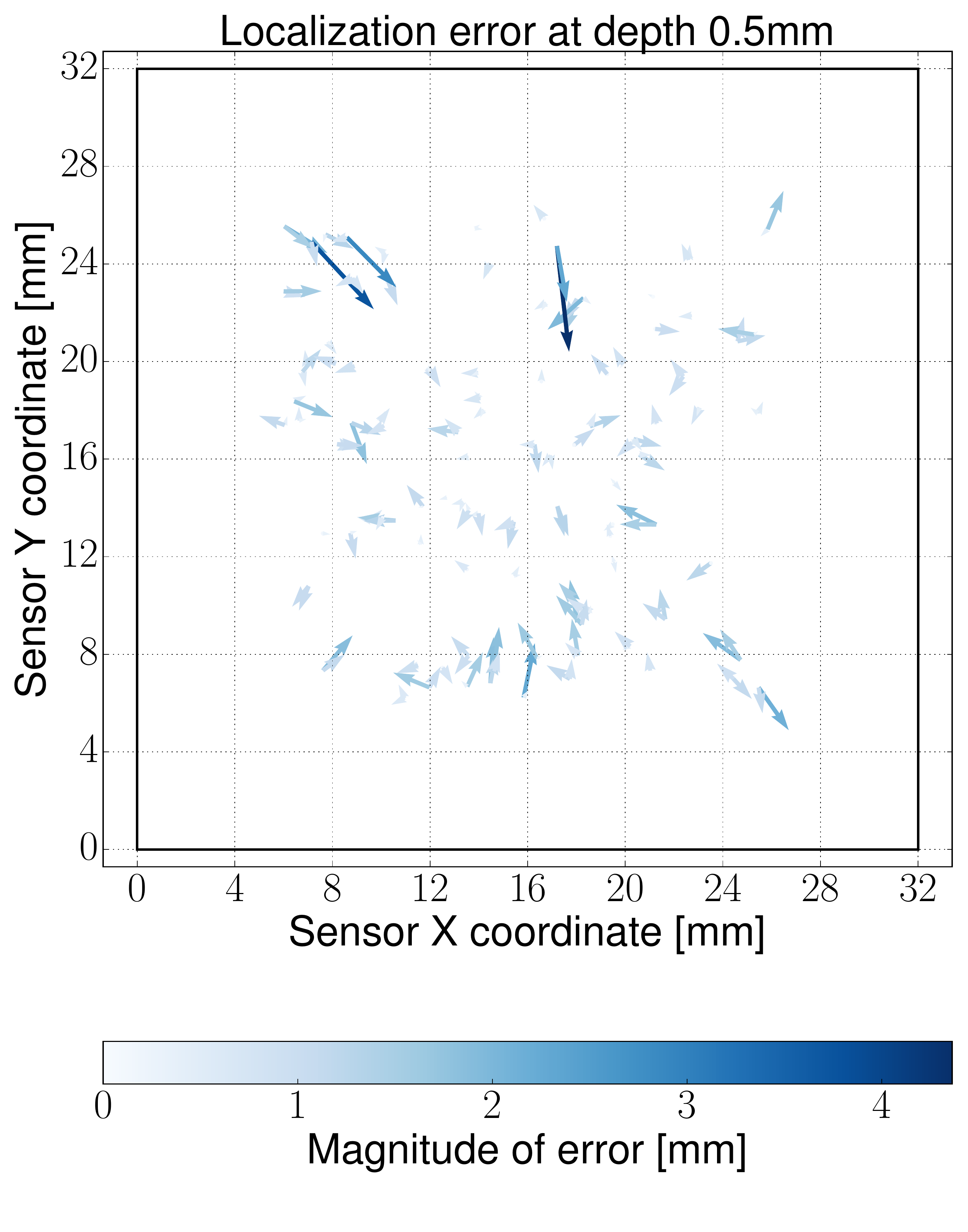}}\hfill
\subfloat[]{\label{ex2}
\includegraphics[clip, trim=0.25cm 1.2cm 0.25cm 0.25cm, width=0.30\textwidth]{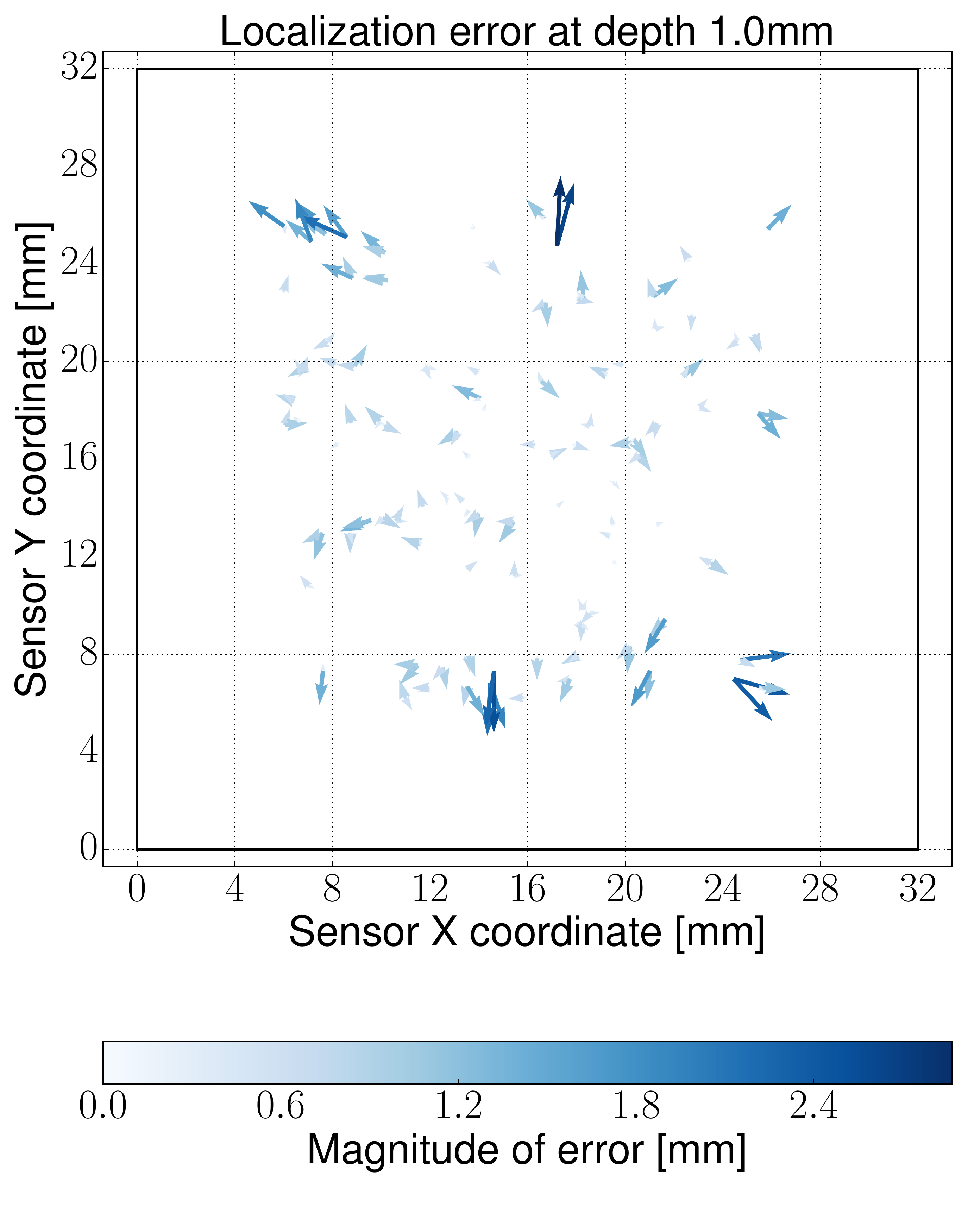}}\\
\subfloat[]{\label{ex3}
\includegraphics[clip, trim=0.25cm 1.2cm 0.25cm 0.25cm, width=0.30\textwidth]{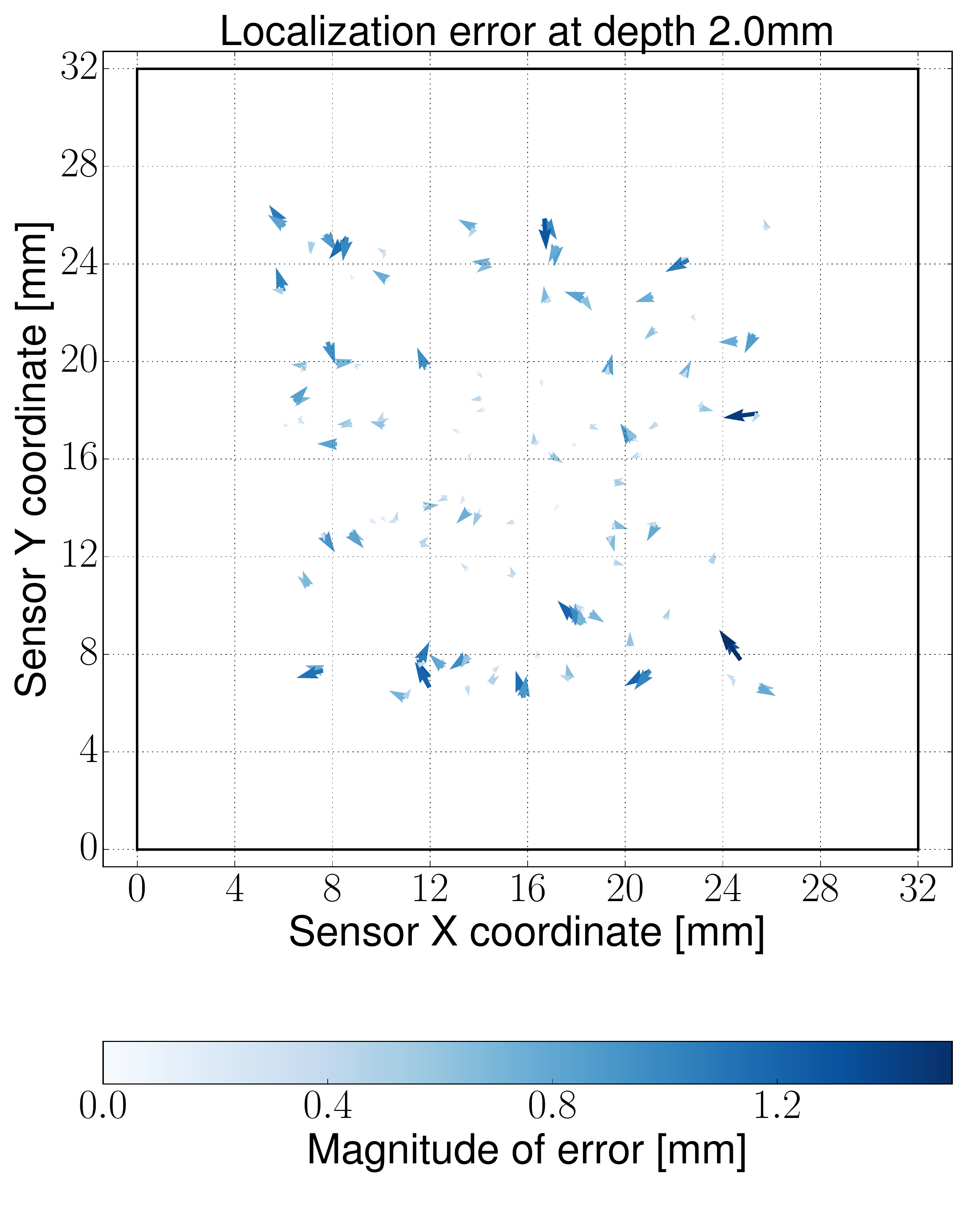}}\hfill
\subfloat[]{\label{ex4}
\includegraphics[clip, trim=0.25cm 1.2cm 0.25cm 0.25cm, width=0.30\textwidth]{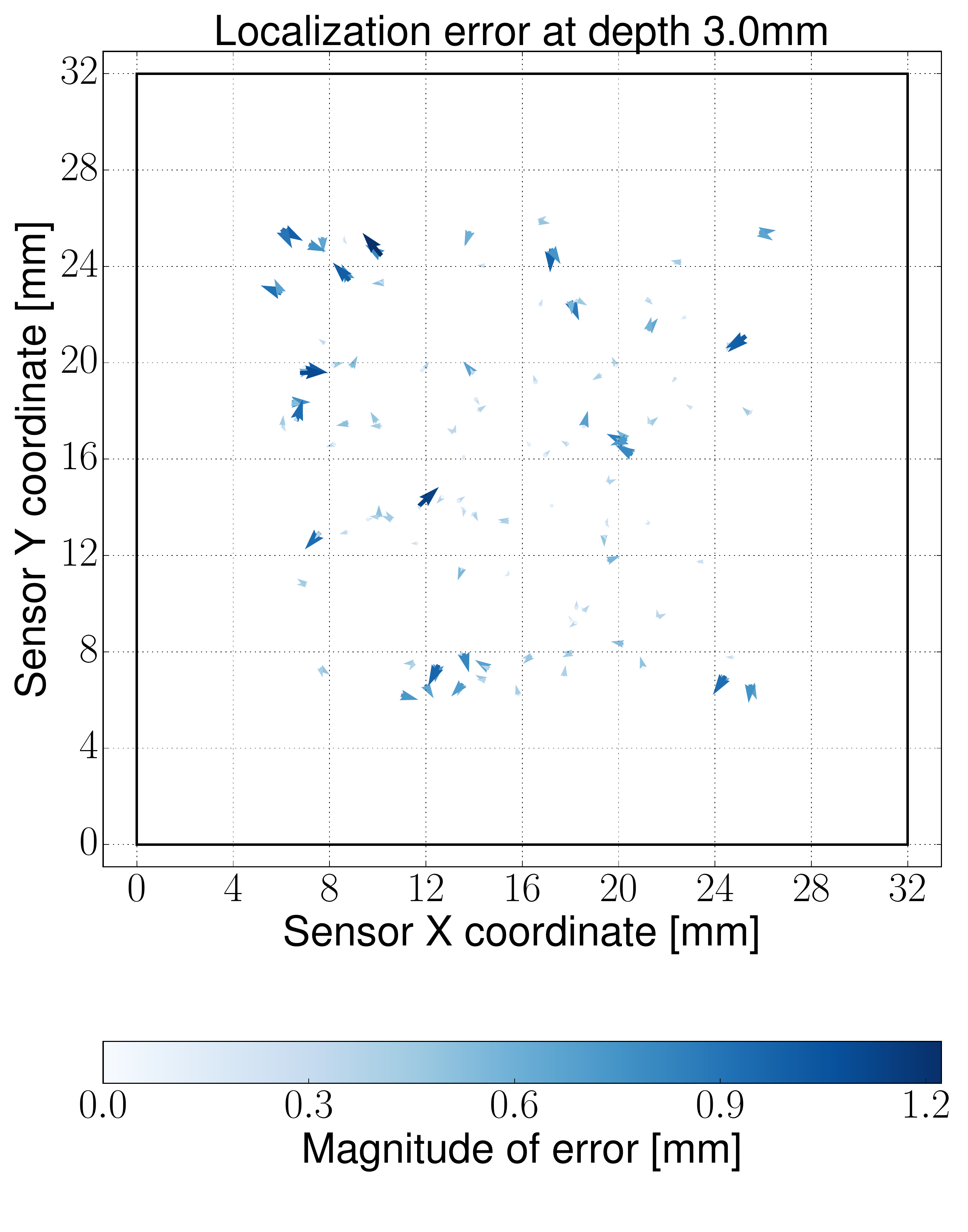}}\hfill
\subfloat[]{\label{ex5}
\includegraphics[clip, trim=0.25cm 1.2cm 0.25cm 0.25cm, width=0.30\textwidth]{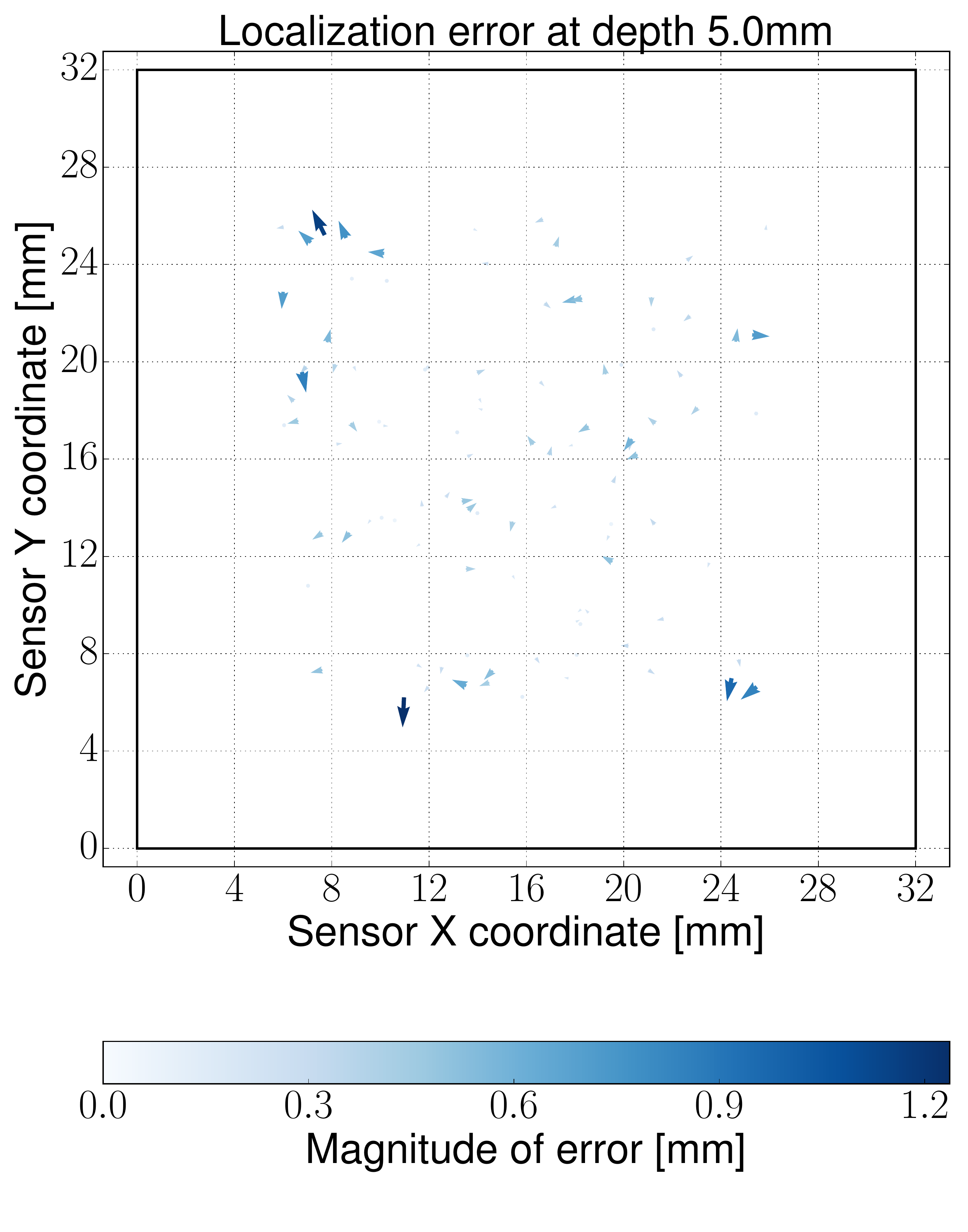}}

\caption{Localization results for ambient light test dataset. Each arrow represents one indentation in our test set; the base is at the ground truth location while the tip of the arrow shows the predicted location.}
\label{fig:Light_results}
\end{figure*}

Regression results for localization on the light dataset are presented
on Fig.~\ref{fig:Light_results} for a few representative depths. At
depth 0.1mm the signals are still not good enough to provide accurate
localization, but as we indent further down localization improves well
beyond sub-millimeter accuracy.

The regression for depth can only be visualized for a specific
location. Fig.~\ref{fig:depth} shows the
performance at three different locations in our ambient light test dataset. Since this regressor is trained only on contact data, it is not
able to predict negative depths which causes error to be greater at
depths close to 0mm. At 0.5mm and deeper, the depth prediction shows
an accuracy of below half a millimeter.

Accuracy in localization and depth for both of our test datasets is presented in more detail in Tables ~\ref{light_table} and
~\ref{dark_table}.

\begin{table*}[ph]
\centering
\caption{Localization and depth accuracy for ambient light dataset}
\label{light_table}
\begin{tabular}{lcccccc}
\hline
\\[-3mm]
\multicolumn{1}{c|}{\textbf{Depth}} & \multicolumn{3}{|c}{\textbf{Localization Accuracy}}                      & \multicolumn{3}{|c}{\textbf{Depth Accuracy}}                             \\
\multicolumn{1}{c|}{}               & \multicolumn{1}{c}{\textbf{Median Err.}} &  \multicolumn{1}{c}{\textbf{Mean Err.}} & \multicolumn{1}{c|}{\textbf{Std. Dev}}  & \multicolumn{1}{c}{\textbf{Median Err.}} &  \multicolumn{1}{c}{\textbf{Mean Err.}} & \multicolumn{1}{c}{\textbf{Std. Dev}} \\ \hline
\\[-3mm] \hline
\\[-2mm]

0.1 mm         & 2.189                & 2.369              & 1.508              & 0.167                & 0.167              & 0.079              \\
0.5 mm         & 0.717                & 0.838              & 0.616              & 0.050                & 0.060              & 0.048              \\
1.0 mm         & 0.641                & 0.744              & 0.530              & 0.048                & 0.065              & 0.051              \\
2.0 mm         & 0.421                & 0.500              & 0.316              & 0.040                & 0.048              & 0.037              \\
3.0 mm         & 0.362                & 0.413              & 0.256              & 0.037                & 0.047              & 0.041              \\
5.0 mm         & 0.314                & 0.359              & 0.222              & 0.081                & 0.092              & 0.062             
\end{tabular}
\end{table*}

\begin{table*}[ph]
\centering
\caption{Localization and depth accuracy for dark dataset}
\label{dark_table}
\begin{tabular}{lcccccc}
\hline
\\[-3mm]
\multicolumn{1}{c|}{\textbf{Depth}} & \multicolumn{3}{|c}{\textbf{Localization Accuracy}}                      & \multicolumn{3}{|c}{\textbf{Depth Accuracy}}                             \\
\multicolumn{1}{c|}{}               & \multicolumn{1}{c}{\textbf{Median Err.}} &  \multicolumn{1}{c}{\textbf{Mean Err.}} & \multicolumn{1}{c|}{\textbf{Std. Dev}}  & \multicolumn{1}{c}{\textbf{Median Err.}} &  \multicolumn{1}{c}{\textbf{Mean Err.}} & \multicolumn{1}{c}{\textbf{Std. Dev}} \\ \hline
\\[-3mm] \hline
\\[-2mm]
0.1 mm         & 3.502                & 3.832              & 2.202              & 0.182                & 0.183              & 0.072              \\
0.5 mm         & 1.102                & 1.333              & 0.985              & 0.050                & 0.058              & 0.044              \\
1.0 mm         & 0.779                & 0.840              & 0.535              & 0.056                & 0.064              & 0.047              \\
2.0 mm         & 0.547                & 0.657              & 0.452              & 0.047                & 0.056              & 0.046              \\
3.0 mm         & 0.410                & 0.487              & 0.322              & 0.039                & 0.050              & 0.042              \\
5.0 mm         & 0.316                & 0.392              & 0.288              & 0.089                & 0.109              & 0.072             
\end{tabular}
\end{table*}

\section{Conclusions}

In this paper, we explore the use of light transport through a clear
medium as the core transduction method for a tactile pad. We
develop touch sensors with multiple light emitter and
receivers around the perimeter of a clear elastomer pad. By measuring
light transport between each emitter-receiver pair we collect a very
rich signal set characterizing indentation performed anywhere on the
sensor surface. We then mine this data to learn the mapping between
our signals and the variables characterizing indentation. The result
is a low-cost, easy to manufacture tactile pad that displays many
desirable characteristics: sensitivity to initial contact, and
sub-millimeter accuracy in predicting both indentation location and
depth throughout most of the operating range.

Our sensor achieves sub-millimeter accuracy over a $400
mm^2$ workspace, while reporting indentation depth within $0.5mm$
throughout most of the operating range. This sensitivity is the result
of leveraging two different modes in which the indenter affects light transport in the PDMS: the first mode being capable of
detecting small indentations during initial contact, and the
second mode detecting larger indentations for stronger
contact. We use depth here as a proxy for indentation force, based on
a known stiffness curve for our sensor; it is also possible to adjust
the PDMS layer thickness and stiffness. A stiffer PDMS layer will require a larger force to activate the second sensing mode.

Future work will be focused on adapting this method for
arbitrary geometries, such that any surface can be covered with this
method. There is also the possibility of learning additional variables, such as shear forces or torsional friction. Other
important factors were also left out of this preliminary study, such as an
analysis of temporary properties, especially regarding
signal drift that might arise over long term periods. Repeatability,
hysteresis and sensitivity to environmental factors are also important
metrics that should be further analyzed. Another improvement
could be to incorporate multi-touch capability into our detection
algorithms.

We believe that ultimately the number of variables that can be learned
and how accurately we can determine those variables depends on the raw
data that can be harvested from the sensor. Increasing the number of
sensing units or even incorporating different sensors embedded into
the elastomer can extend the sensing modalities in our
sensor. Different learning methods like deep neural networks might
also improve the performance or capability of the sensor. All these
ideas will be explored in future work.

\bibliographystyle{IEEEtran}
\bibliography{bib/tactile,bib/grasping,bib/thesis}

\end{document}